# CNN Based Posture-Free Hand Detection

Richard Adiguna, Yustinus Eko Soelistio
Information System Dept.
Universitas Multimedia Nusantara
Indonesia
Email: {richard.adiguna*, yustinus.eko}@{student*}.umn.ac.id

*Abstract*—Although many studies suggest high performance hand detection methods, those methods are likely to be overfitting. Fortunately, the Convolution Neural Network (CNN) based approach provides a better way that is less sensitive to translation and hand poses. However the CNN approach is complex and can increase computational time, which at the end reduce its effectiveness on a system where the speed is essential. In this study we propose a shallow CNN network which is fast, and insensitive to translation and hand poses. It is tested on two different domains of hand datasets, and performs in relatively comparable performance and faster than the other state-of-the-art hand CNN-based hand detection method. Our evaluation shows that the proposed shallow CNN network performs at 93.9% accuracy and reaches much faster speed than its competitors.

## I. Introduction

Advancement in hand detection methods have enable humans and computer to interact with one another in a complex and personal way [1-6]. Giving ability for computer to detect hand will give new experience in interacting with computer in a more natural and flexible way. Hand detection has becoming an important area of research because of it has wide applications for human computer interaction [1]. However, designing a detection and recognition system to a computer is quite difficult because the complexity of background and high variability in the hand postures. There are already several methods suggested to handle these problems such as [2-6].

One method is proposed by [2] who utilize thermal sensors to detect and track the hand position. Their detection method works by comparing the binary thermal images and the thermal template. The hand is detected in the agreement between the binary image and the template. This method can be used in any illumination condition. However, the use of specialized device (thermal sensor) makes the method less usable in everyday situations.

Others methods take a different approach by implementing appearance based technique [3-6] which can be used in everyday situation (e.g. using webcam or standard camera). This ability makes the appearance based technique an interesting approach for hand detection methods. Here we describe four appearance based hand detection methods that represent the common approaches in the literatures.

The first method [3] utilize 2D Fourier transformation to convert the hand images into 2D Fourier images. Similar to the method by [2], this method also works by comparing the input (after converted to 2D Fourier image) to a template. The method assumes six hand postures' templates, thus limit its usability to classify only the six postures.

Another approach is proposed by [4] who exploit the unique shape and skin tone features the hand. Their method uses boosted classifier tree for hand detection. The structure of the tree consist of a general hand detector at the top and a more specific hand detector at lower level. The hand detector is applied using k-medoid clustering algorithm on hand shape descriptor. The descriptor takes form of edge histogram and skin model. The hand will be detected when descriptor is close (in Euclidean distance) to the hand samples (i.e. the hand templates in the k-medoid cluster).

Similarly, [5] implement hand shape and colors detector in their method. However, they add additional features of context to provide extra clues for hand detection (i.e. that the end of the arm is more visible than the hand so that detecting the end of the arm will give a better prediction of the location of the hand). Their method works in two steps: features detection and classifier. The features detection consist of: (1) the hand's rough location by comparing the hand and the face skin tone (the hand and the face should have similar color), (2) the hand's context detection by using deformed model, and (3) the hand's shape detector using HOG features extraction. The three features are then classified using linear SVM classifier.

The last approach utilize the structure of (CNN) deep learning as classifier [6]. Unlike its predecessor that prone to overfitting since their reliance on templates, this CNN based approach should be more insensitive to translation and hand postures. The CNN is a deep structure consist of four parts: (1) shared network (3 convolution + ReLu + pooling layers), (2) rotation network (2 convolution + pooling + ReLu layers, followed by 3 fully connected layers), (3) derotation layer which rotate the probable hand image to predetermine angle, and (4) detection network (2 convolution + pooling + ReLu + pooling layers, followed by 3 fully connected layers). The first two networks responsible to predict the rotation of the probable hand image, and produce the rotation estimation of the image. The estimation is fed to the derotation layer where it calculate the new in-plane rotation angle for all the pixels in the image. The new rotated image is then used by the detection network to classify the hand / no-hand condition. This CNN approach is considered superior than the four aforementioned approaches

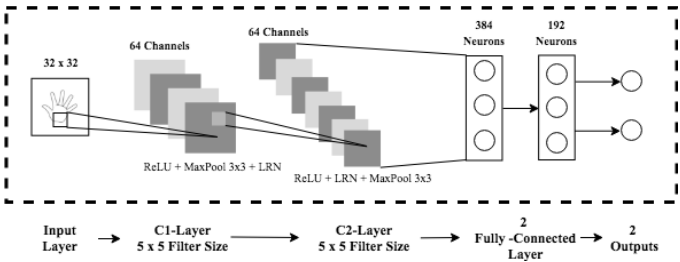

Fig. 1. Convolutional Neural Network Structure based on [7].

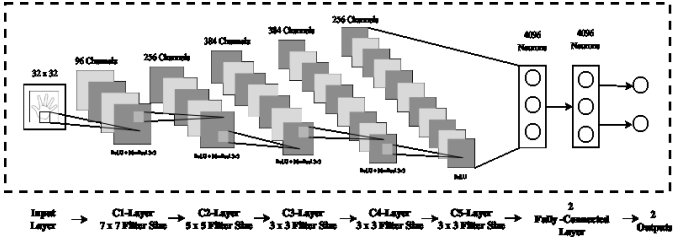

Fig. 2. Convolutional Neural Network Structure based on [8].

since CNN provides the mean to handle large interclass variations.

The method [6] constitute the state-of-the-art hand detection method at the moment. In spite of its high performance in detecting the hand in various postures, we argue over its complexity on simple binary classification task. They reported a high computation time (8 seconds / image) which reflects the high computation cost of their method. A high computation time should limit the usefulness of this method when detection speed is essential (i.e. real-time system). Therefore there is still a need for a fast high performance hand detection method.

This study proposes a fast and accurate CNN-based hand detection method to tackle the high computational time suffered by theaforementioned state-of-the-art method. The CNN structure is based on existing CUDA-Convnet2structure [7] and Chen [8]. The structure is evaluated on two different NUS datasets [1]. By comparison, we empirically proved that our implementation ofCUDA-Convnet2structure [7] can be generalized better than the one of [8] and it is faster than [8] and [6].

The rest of the article will be presented in the following order. The dataset collection will be discuss in the section II, our methodology in section III, and show our result in section IV. Section V will be a discussion and the comparison with the previous work and section VI will be the conclusion of this paper.

## II. PROPOSED METHOD

Our proposed method is based on the CUDA-Convnet2 network's structure proposed by [7]. The structure consists of two convolutions layers with pooling, ReLu, and normalization functions, then followed by two fully connected layers (Figure 1).

The network is trained on NUS II dataset, and compared with similar but deeper network from [8]. The network from [8]

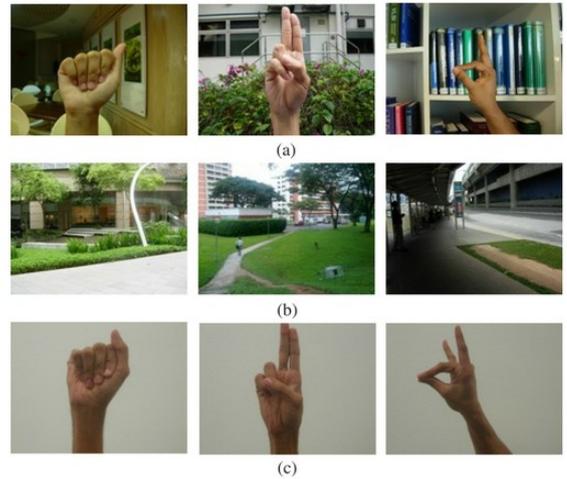

Fig. 3. (a) NUS II hand posture with background images (b) NUS II background images (c) NUS I hand posture images.

consists of five convolution layers, with ReLu and pool, followed by two fully connected layers (Figure 2).

## III. IMPLEMENTATION AND DATA PREPARATION

We implements the networks of [7] and [8] as closely as possible as reported in the reports and project's site. However there are some hyperparameters that are missing, so we make some customization on the networks. There are eight hyperparameters that might be different from the original networks: (1) learning rate ($10^{-4}$), (2) learning rate decay by a factor of 0.8 for every epoch, (3) dropout ($0.4$), (4) batch size of 32, (5) epoch 15 with iteration of 112, (6) Adam Optimizer as learning functions, (7) all weights are initialized using normal distribution with standard deviation of 0.005, while all bias are set to zero, and (8) softmax function as the last output layer. We also change the input size into $32 \times 32$ (RGB channels) and make the two outputs of one-hot vector (i.e. {0,1} and {1,0} for hand and no-hand condition respectively). Those conditions areidentical for both networks. The implementation use Python version 3.6.0 and Tensorflow version 1.5.

We use both the NUS I [1] and II [9] for training and evaluation. For evaluation we use 10-fold cross validation evaluation method on the NUS II, and positive test evaluation on NUS I.

The images in the two datasets are in size of $120 \times 160$ pixels. The NUS I hand posture dataset consists of 480 images with 10 classes of hand posture (240 in BW and 240 in RGB channels). We only use the RGB images for evaluation. The NUS II hand posture dataset consists of 2750 images on ten hand postures with two kinds of noises (human and background). We only use the images with background noises for training and evaluation (2000 images).

The selected hand images from NUS II are merged with 2000 no-hand images provided in the NUS II dataset. Combining the two sets, we implement the 10-fold cross validation which makes 3600 images (equal size of hand and no-hand conditions) for training and 400 images for evaluation

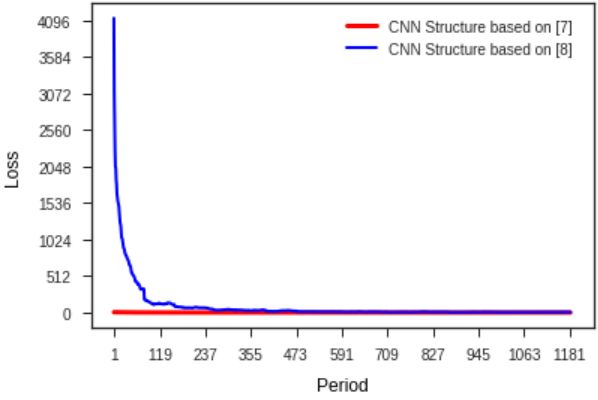

Fig. 4. Loss from training process of our implementation of [7] and [8].

purposes. Some examples of images in our selected dataset are shown in Figure 3.

To evaluate the generalization of the networks on different data domain, we evaluate the network (trained in each fold) on the NUS I RGB images.

## IV. TRAINING AND EVALUATION

The training of both networks are conducted on NUS II dataset using 10-fold cross validation method. Each fold is trained for 1680 iterations.

Figure 4 shows the loss function on each iteration of the training. Visual inspection shows a similar training behavior at the end of both networks. Our implementation of [8] start the training worse than the one from [7] until about the first 120 iterations. The training slops from the two networks merge after the 240 iterations. Both networks shows a small but linearly decreasing slope. The decreasing slopes indicate a good enough learning process of the networks.

The evaluations of the networks are performed in terms of performance (classifier's accuracy) and speed. We compare the performance of the two networks and other, and make a comparative TFlops evaluation on the speed from both network and [6]. The results of the evaluations are presented in the next two subsections.

### A. Performance Evaluation

We compare the performance of our implementation of [7] and [8]. We implement two types of evaluation, (1) 10-fold cross validation method on NUS II dataset, and (2) positive test on NUS I dataset. The evaluations compare the mean accuracy ($\mu$) and standard deviation ($\sigma$).

In the first evaluation, the networks from [7] and [8] achieve $\mu_{[7]} = 93.9\%$ and $\mu_{[8]} = 97.5\%$, with $\sigma_{[7]} = 1.6\%$ and $\sigma_{[8]} = 3.9\%$. The structure from [8] achieve a better performance over the [7]. However when evaluated on dataset from different domain (the positive test in the second evaluation), the [7]'s structure performs better than the [8]'s ($\mu_{[7]} = 83.2\%$, $\mu_{[8]} = 73.8\%$). This result suggests that the structure from [7] can be better generalized then the [8] for hand detection tasks.

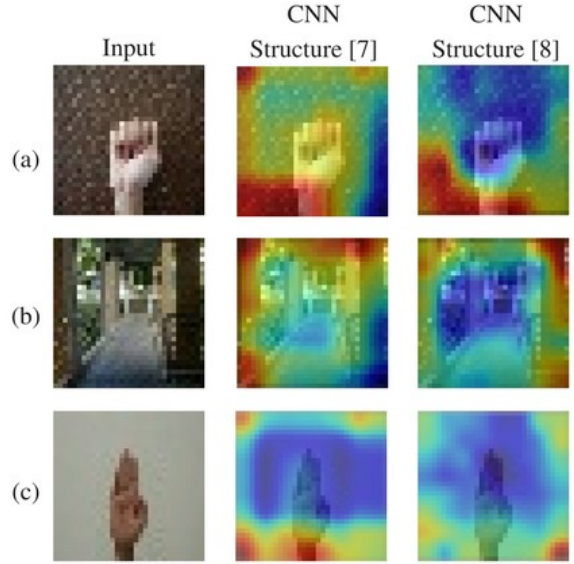

Fig. 5. Convolution results from the last layer of the two CNN structures. (a) hand images from NUS II dataset, (b) no-hand images from NUS II dataset, (c) hand posture images from NUS I dataset.

### B. Speed Evaluation

Speed evaluation is performed by relative comparison. We compare the average computational speed for classifying one image and normalize the speed by the ratio of the TFlops of [6] and our computational power. We use intel i7-4700HQ 2.4GHz. The [6] uses a quad core 2.9 GHz computer with Nvidia Titan X GPU.

The [6] reports the speed of 8 seconds-per-image for classification. However, [6] does a sequential search using linear SVM on a $500 \times 600$ pixels space for hand region checking so the true classification speed should be much faster than the one reported. We took the best case of classification speed which can be calculated by dividing the reported speed with the hand region search space, thus 0.03 ms is the estimated best-case time needed to classify one image on their computer.

Our experiments show computation time of 4.31 ms for [7]-based and 8.64 ms for [8]-based structures.

Based on [12-13], our computer can perform on 26.5 GFlops while theirs on 11 TFlops which makes the ratio 1:415 computational power between our system and theirs.However since we only use linear CPU implementation and [6] utilize a 3072 cores GPU then the ratio can reach 1:1274880. Using this ratio, we calculate the computational time between our structures and [6] which are 4.31 ms on ours [7] and 38246.4 ms for the [6]. By this comparison, our network is much faster by a fraction of 8873.87.

## V. DISCUSSION

The evaluations show that the structure from [7] can be used for hand detection task. It is insensitive to hand pose and performs between 83.2% to 93.9% accuracies. Although it performs less than the one proposed by [6], it can be generalize with good enough accuracy. The effectiveness of the method appears to be caused of the better (than the [8]) localization of the wrist and the hand.

To see what actually CNN kernels do, Figure 5 present the convolution result from the last linear combination between kernels and current input from CNN structure based on [7] and CNN structure based on [8]. For the hand image processed by the [7]'s structure, the high intensity region is at wrist area. Although, the intensity at the area of hand is not relatively high but there is an increment of intensity around the hand area. On the other hand, the intensity of the no-hand images appears random which might just be some noises. This behavior is less obvious on the [8]'s structure.

The network ([7]-based) performs faster than the [8] and the [6]. By comparison, our chosen network can reach up to 8873 times faster than the [6]. Based on the test, the network can be used on 161 fps system which is much above the average standard camera (30 fps).

## VI. CONCLUSION

In this study we propose a fast and accurate hand detection method. It is proven to be generalizable and insensitive to hand pose which extend its usefulness on everyday usage. Although, as any other methods, still suffer the tradeoff between the speed and performance, we believe this method is an attractive alternative for hand detection method.

ACKNOWLEDGMENT

The result in this study is part of the theses by Richard Adiguna, and supported by LPPM dept. and Big Data Lab. of Universitas Multimedia Nusantara.